\documentclass[10pt,twocolumn,letterpaper]{article}
\usepackage{tikz}
\usepackage{pgfplots}
\usepackage{placeins}
\usepackage{float}
\usepackage{graphicx}
\usepackage{caption}
\usepackage{booktabs}
\usepackage{dblfloatfix} % better placement for figure*/table* (top/bottom)
\usepackage{cuted} 

\newcommand{\modelname}{TriDiff-4D\xspace}

\usepackage[pagenumbers]{wacv} % To force page numbers, e.g. for an arXiv version

\definecolor{wacvblue}{rgb}{0.21,0.49,0.74}
\usepackage[pagebackref,breaklinks,colorlinks,allcolors=wacvblue]{hyperref}

\def\wacvPaperID{2460} % *** Enter the WACV Paper ID here
\def\confName{WACV}
\def\confYear{2026}

\title{TriDiff-4D: Fast 4D Generation through Diffusion-based Triplane Re-posing}
\author{
  Eddie Sheung\textsuperscript{1}, \quad 
  \hspace{-1ex}Qihao Liu\textsuperscript{1}, \quad
  \hspace{-1ex}Wufei Ma\textsuperscript{1}, \quad 
  \hspace{-1ex}Prakhar Kaushik\textsuperscript{1}, \quad
  \hspace{-1ex}Jianwen Xie\textsuperscript{2}, \quad
  \hspace{-1ex}Alan Yuille\textsuperscript{1}
  \vspace{4px}
    \\
  \vspace{4px}
  \hspace{-1.5ex}\textsuperscript{1} Johns Hopkins University \qquad
   \hspace{-1.5ex}\textsuperscript{2} Lambda Inc 
    \\
    \vspace{4px}
}
\usepackage{afterpage}  % Add to preamble

\begin{document}
\maketitle
% \begin{figure*}[!t]
%     \centering
%     \vspace{-13mm}
%     \includegraphics[width=\linewidth]{imgs/prewview2.png}
%     \vspace{-13mm}
%     \caption{\modelname is a novel 4D generative pipeline that enables high-quality, controllable 4D avatar generation from text using diffusion-based triplane re-posing. By explicitly modeling 3D structure and motion priors within the diffusion model, learned from large-scale 3D and motion datasets, it produces anatomically accurate, motion-consistent, dynamic, and visually coherent  avatars that generate 14 frames of 3D object sequences in just 36 seconds on a single H100 GPU. 
%     }
%     \label{fig:teaser}
% \end{figure*}

\begin{strip}
\vspace{-8mm} % More space from title
\centering
\includegraphics[width=0.8\textwidth]{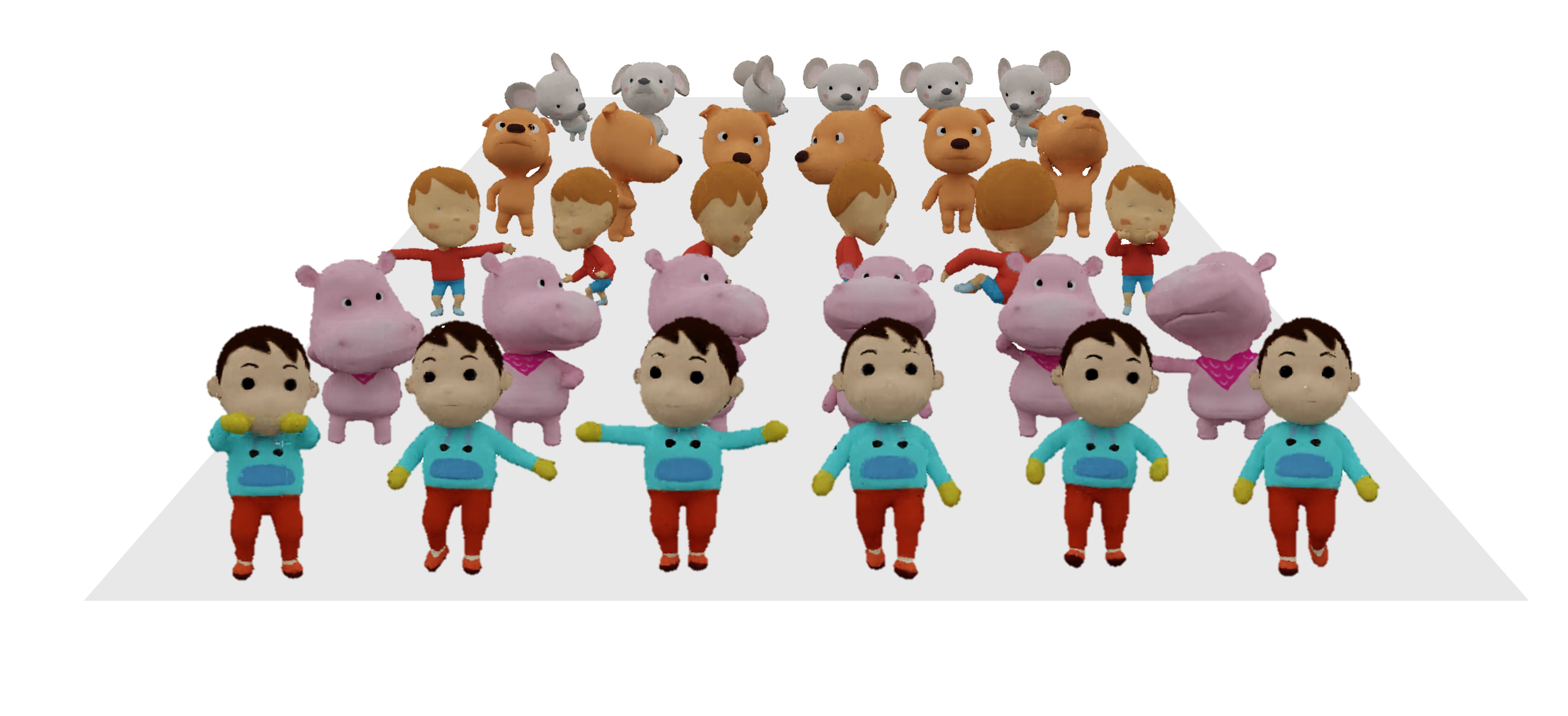}
\vspace{-2mm} % Space between image and caption
\captionof{figure}{\modelname is a novel 4D generative pipeline that enables high-quality, controllable 4D avatar generation from text using diffusion-based triplane re-posing. By explicitly modeling 3D structure and motion priors within the diffusion model, learned from large-scale 3D and motion datasets, it produces anatomically accurate, motion-consistent, dynamic, and visually coherent avatars that generate 14 frames of 3D object sequences in just 36 seconds on a single H100 GPU.}
\label{fig:teaser}
% \vspace{2mm} 
\end{strip}

\begin{abstract}
With the increasing demand for 3D animation, generating high-fidelity, controllable 4D avatars from textual descriptions remains a significant challenge. 
Despite notable efforts in 4D generative modeling, existing methods exhibit fundamental limitations that impede their broader applicability, including temporal and geometric inconsistencies, perceptual artifacts, motion irregularities, high computational costs, and limited control over dynamics.
To address these challenges, we propose \modelname, a novel 4D generative pipeline that employs diffusion-based triplane re-posing to produce high-quality, temporally coherent 4D avatars. 
Our model adopts an auto-regressive strategy to generate 4D sequences of arbitrary length, synthesizing each 3D frame with a single diffusion process. 
By explicitly learning 3D structure and motion priors from large-scale 3D and motion datasets, \modelname enables skeleton-driven 4D generation that excels in temporal consistency, motion accuracy, computational efficiency, and visual fidelity. Specifically, \modelname first generates a canonical 3D avatar and a corresponding motion sequence from a text prompt, then uses a second diffusion model to animate the avatar according to the motion sequence, supporting arbitrarily long 4D generation.
Experimental results demonstrate that \modelname significantly outperforms existing methods, reducing generation time from hours to seconds by eliminating the optimization process, while substantially improving the generation of complex motions with high-fidelity appearance and accurate 3D geometry.

\end{abstract}

\section{Introduction}
\label{sec:intro}

\begin{figure*}
    \centering
    \includegraphics[width=\linewidth]{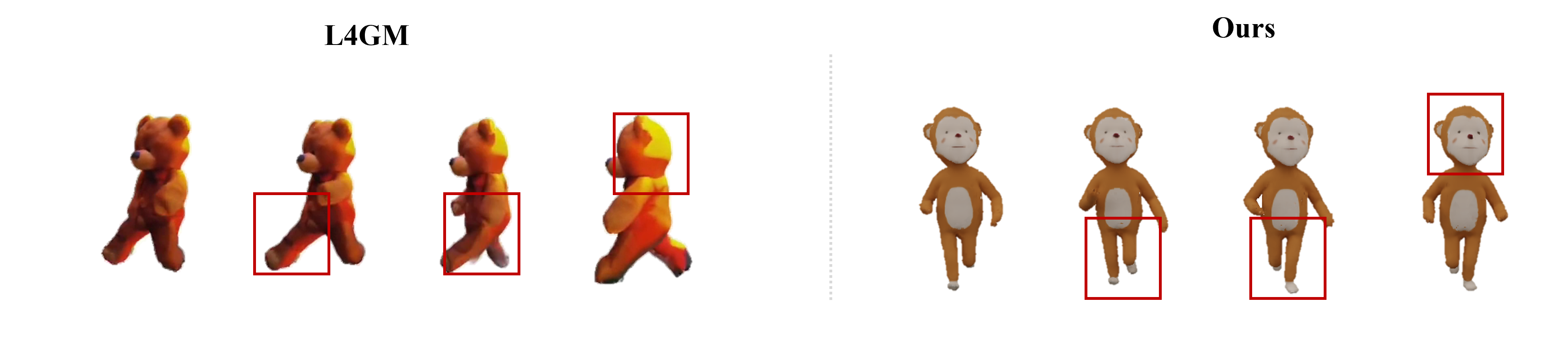}
    \caption{\textbf{Running motion comparison~\cite{ren2024l4gm} (left) and our method (right)}. The baseline exhibits unrealistic geometric stretching, particularly evident in limb elongation during dynamic movements, while our method maintains consistent proportional geometry and structural integrity throughout the motion sequence. 
    }
    % \eddie{This is the most similar visualization I can find, but they uses shortage is at the sides and back. Compare this with our front walking is wired? }
    \label{fig:Jelly_effect}
\end{figure*}

%% LQH: this paragraph only talk about challenges in 3D generaion
The demand for realistic, expressive, and controllable 3D avatars is rapidly growing across a range of domains, including gaming, and virtual and augmented reality (VR/AR). 
However, creating digital avatars that combine high visual fidelity with fine-grained control over movement and pose remains a complex challenge. 
Existing 3D/4D generation methods often depend on (multi-view) 2D priors ~\cite{xie2025sv4ddynamic3dcontent}~\cite{zeng2024stag4dspatialtemporalanchoredgenerative}~\cite{zheng2024unifiedapproachtextimageguided}, leading to a difficult trade-off between appearance quality and 3D geometric accuracy. 
A well-known issue in such approaches is the Janus problem~\cite{poole2022dreamfusion}, where geometric inconsistencies across viewpoints can cause artifacts such as characters displaying multiple faces when viewed from different angles.
%% challenges in 4D generation
Building on the challenges of 3D generation, 4D generation introduces additional complexity by requiring 3D content to evolve coherently over time, demanding high motion realism and temporal consistency.
For example, due to the inherent difficulty of synchronizing motion patterns with structural deformations, existing methods often suffer from temporal inconsistencies and unrealistic non-rigid deformations, commonly referred to as the ``jelly effect'' (as illustrated in Fig.~\ref{fig:Jelly_effect})~\cite{ren2024l4gmlarge4dgaussian}. 
Moreover, achieving realistic motion and visual detail remains difficult, as it requires balancing geometric consistency and texture fidelity across frames~\cite{bahmani20244d}.

%% To solve these problems, current methods and limitations
To resolve these fundamental difficulties, current 4D generation methods often rely on complex intermediate representations such as detailed meshes~\cite{SMPL:2015} or keypoint structures~\cite{chai2024star}, which are time-consuming and difficult to obtain and manipulate. 
A substantial body of research leverages pre-trained image or video diffusion models as priors to guide 4D content creation~\cite{singer2023text4d}. 
They typically employ Score Distillation Sampling (SDS) to optimize various 4D representations, including NeRF~\cite{mildenhall2021nerf}, 3D Gaussian Splatting~\cite{tang2024dreamgaussian}, or deformable meshes, as seen in methods like 4D-fy~\cite{bahmani20244d} and 4Dynamic~\cite{yuan20244dynamic}.
However, they are usually constrained to low-resolution outputs and simple motions, primarily due to the quality of 2D generations, computational inefficiencies, and the scarcity of high-quality 4D training data~\cite{lin2023magic3d}. Notably, ``Jelly effect'' and ``Janus problem'' persist as major challenges in these approaches, significantly degrading the animation quality and limiting their practical applicability. 

% and continue to suffer from issues such as the ``Janus problem'' and ``Jelly effect''.

%% our solution
To address these limitations, we propose \modelname, a \textbf{Tri}plane Re-posing \textbf{Diff}usion model for \textbf{4D} generation.
Instead of relying on computationally expensive optimization loops with 2D image and video priors using SDS loss, \modelname efficiently encodes 3D object and motion knowledge, and reduces the generation time from hours to seconds.
In addition, our model addresses key limitations of existing methods by combining an efficient triplane representation with direct diffusion-based reposing. This integration ensures frame-to-frame temporal coherence, eliminates the Janus problem across viewpoints, and preserves anatomical correctness throughout complex motion sequences. 
Furthermore, unlike approaches that rely on mesh deformation and video-to-4D techniques, we condition directly on 3D skeletons within the triplane feature space. This shift in conditioning dimensionality enables us to maintain volumetric consistency across all viewpoints, effectively eliminating the jelly-like wobbling effect.

%Our approach proceeds through the following stages: 

Specifically, our proposed \modelname begins by generating an initial static 3D avatar represented by triplane features, directly derived from a textual description specifying the object's category and appearance. 
Simultaneously, a sequence of skeleton poses is generated from a separate textual description specifying the desired motion with a text-to-motion models, renowned for generating expressive and complex motions. 
After that, a novel diffusion-based re-posing mechanism is used to generate a sequence of animated triplane features. 
More precisely, this module takes the initial avatar's triplane features and the generated skeleton sequence as conditional inputs, iteratively transforming the pose information within the latent space for each frame. 
This ensures that the avatar accurately aligns with each target skeleton pose while maintaining appearance consistency throughout the motion. 
Finally, the re-posed sequence of triplane features are decoded into a dynamic, temporally smooth, and view-consistent 4D avatar. 
Our framework supports flexible rendering options, with compatibility for both Neural Radiance Fields (NeRF) and Gaussian Splatting models as decoders, allowing users to select the optimal rendering approach based on their specific requirements.

To demonstrate its effectiveness, we compare \modelname with leading 4D avatar generation methods across generation speed, qualitative comparisons, and user preference study. The results demonstrate that our method delivers superior visual quality, natural motion, accurate geometry, and overall appeal, while being significantly more computationally efficient than existing approaches. Quantitatively, \modelname achieves a user preference score of 79.59\% versus 20.41\% for the previous open-source state-of-the-art model~\cite{ren2023dreamgaussian4d}, and reduces generation time from 10 minutes to just 0.6 minutes.

In summary, our main contributions are as follows:
\begin{itemize}
    % \item We propose a cohesive text-to-4D generation pipeline, named \modelname, that directly generate high-quality 3D objects with realistic motions in a single forward pass
    % \item A novel diffusion-based methodology for re-posing 3D avatars by modifying their triplane features based on conditioned target skeleton sequences.
    % \item We address  
    
    \item We propose \modelname, a cohesive text-to-4D generation pipeline that directly synthesizes high-quality 3D avatars with realistic motion in a single forward pass within seconds.
    
    \item We introduce a novel diffusion-based model for re-posing 3D avatars by modifying triplane features based on generated skeleton sequences, ensuring temporal coherence and anatomical accuracy across complex motions.
    
    \item We learn 3D structure and motion priors directly from large-scale assets and address key limitations of existing methods by explicitly conditioning 4D generation on our generated 3D avatars and skeletal motion. This approach eliminates view inconsistencies, temporal flickering, and the `jelly-like wobbling effect' commonly seen in competing methods.
    
    % \wufei{Be more specific: One sentence highlighting technical issues our method address and that our method generates ``high-fidelity, view-consistent temporal coherence across 4D avatar outputs''.}
\end{itemize}

% \begin{figure*}
%     \centering
%     \includegraphics[width=\linewidth]{imgs/Jelly_effect.png}
%     \caption{\textbf{Jelly effect comparison between Align Your Gaussians~\cite{ling2024align} (left) and our method (right)}. The baseline exhibits unrealistic deformations, resulting in a jelly-like wobbling effect, while our method preserves consistent geometry and structure. Note: Both depict a ``running'' motion. Frames for the baseline were extracted from its project video due to lack of official code.
%     }
%     \eddie{change the comparison to the same species }
%     \label{fig:Jelly_effect}
% \end{figure*}

\section{Related Work}

\noindent\textbf{Text-to-3D Generation.}
The generation of static 3D-objects from textual prompts has seen significant progress. Direct-3D~\cite{liu2024direct} addresses critical challenges in text-to-3D generation by introducing a triplane feature generation approach for 3D model creation, which provided the initial triplane features for our proposed pipeline. Its architecture features a disentangled triplane diffusion module for separate geometry and color generation, followed by a NeRF auto-decoder. Another foundational work, EG3D~\cite{chan2022efficient}, introduced an efficient geometry-aware 3D Generative adversarial Network(GAN) that utilizes a hybrid explicit-implicit triplane architecture, demonstrating the efficacy of triplane for 3D generation. And many other approaches leverage the popular Score Distillation Sampling(SDS), like one of the popular methods DreamFusion~\cite{poole2022dreamfusion} which distill knowledge from pre-trained 2D text-to-image diffusion models into 3D representations. Also, some methods use multi-view generation for high-resolution 3D content creation like LGM~\cite{tang2024lgm}. Overall, the research for 3D generation has shifted towards directly generating comprehensive 3D representations, which prioritizes more efficient and streamlined methodology that enable more direct translation of input descriptions into high-fidelity 3D assets~\cite{liu2023zero}.

\noindent\textbf{Motion Generation and Skeleton-based Animation.}
Early implementations of motion generation rely on diffusion model, such as MDM and SinMDM~\cite{ho2020denoising, tevet2022human, raab2023single}, use transformer-based architecture and operate by denoising from sampled motion sequences. This process allows the model to generate diverse and realistic human motions conditioned on various inputs such as text prompts or action labels. Furthermore, MoMask~\cite{guo2024momask} is a leading framework for text-driven motion generation. It employs a hierarchical quantization scheme to represent human motion as discrete tokens and uses bidirectional transformers for generation, which achieves a high-fidelity result. 

Beyond text-to-motion, there are numerous studies working on general human pose estimation and motion generation from various inputs~\cite{zhang2023adding, hu2024animate, liu2025revision, xu2024magicanimate}. Several methods specifically focus on using skeleton conditioning to control avatar animation. For instance, Animate Anyone~\cite{hu2024animate}, a novel framework that leverages a UNet-based architecture to generate photorealistic human animations. The system takes skeleton sequences as input and produces high-fidelity 2D videos that precisely follow the skeletal motion while preserving the appearance of the reference subject. This approach effectively bridges the gap between motion control and visual rendering, enabling detailed character animation from simple pose sequences. Furthermore, DreamWaltz-G~\cite{huang2024dreamwaltz} integrates skeleton controls from 3D human templates~\cite{jiang2024smplx} into 2D diffusion models using a skeleton-guided score distillation strategy, combined with a hybrid 3D gaussian avatar representation for expressive animation.

\noindent\textbf{4D generation.}
4D generation has been an active research area as the demand for 3D scenes or avatars is increasing. 4D-fy~\cite{bahmani20244d} is a key temporary method that must be considered. It utilizes a hybrid score distillation sampling(SDS)~\cite{chai2024star} , like some 3D-aware text-to-3D generation models~\cite{tang2023stable, poole2022dreamfusion, yan2024flow}, and optimize a 4D radiance field. This approach aims to balance the appearance quality, 3D structure and motions. Another important approach is video sequence based 4D generation such as DreamGaussian4D~\cite{ren2023dreamgaussian4d}
 and 4DGen~\cite{yin20234dgen}~\cite{wang2023prolificdreamerhighfidelitydiversetextto3d}. These frameworks emerged as a notable contribution addressing critical limitations observed in prior 4D content creation pipelines that predominantly relied on text-to-image and text-to-video diffusion models.~\cite{zhao2024animate124animatingimage4d}

\section{Preliminaries}
\label{sec:preliminaries}

\noindent\textbf{Latent Diffusion Model with Triplane Representation.}
Diffusion models~\cite{ho2020denoising} are a class of generative models that learn to transform a simple noise distribution into a complex data distribution through an iterative denoising process. Latent diffusion models~\cite{rombach2022high} move this process from pixel space to the latent space of a Variational Autoencoder~\cite{kingma2013auto}. 
Here, we follow previous work~\cite{liu2024direct} and consider the diffusion process in the triplane Space~\cite{chan2022efficient}.

Specifically, we consider the triplane representation of the geometric and color information, denoted $F_{geo/color, 0}$ of the input data, and refer to it simply as  $F_0$ in this section. In the forward diffusion process, Gaussian noise is incrementally added to $F_0$:
\begin{equation}
    q(F_t | F_{t-1}) = \mathcal{N}\left(F_t; \sqrt{1 - \gamma_t} F_{t-1}, \gamma_t \mathbf{I}\right),
\end{equation}
where $F_t$ represents the noisy triplane representation at time step $t$, and $\gamma_t$ is a predefined noise schedule with $t \in (0, 1)$. 
With reparameterization, this process allows sampling $F_t$ in a closed form:
\begin{equation}
    F_t = \sqrt{\bar{\alpha}_t} F_0 + \sqrt{1 - \bar{\alpha}_t} \, \epsilon,
\end{equation}
where $\bar{\alpha}_t = \prod_{i=1}^t (1 - \gamma_i)$ and $\epsilon \sim \mathcal{N}(0, \mathbf{I})$. The diffusion model, parameterized by $\Theta$, learns to reverse this noising process by taking $F_t$ as input and reconstructing the clean data with the objective:
% Afterwards, the latent diffusion models (parameterized by $\theta$) learn to  reverse this noising process, which takes $z_t$ as input and is optimized to reconstruct the clean data with the following objective function:
\begin{equation}
\mathcal{L} = \left\| \epsilon - \epsilon_\Theta(F_t, t, c) \right\|_2^2,
\end{equation}
where $c$ is a condition to guide the denoising process. 

\section{Methodology}
\begin{figure*}
  \centering
  \includegraphics[width=\textwidth]{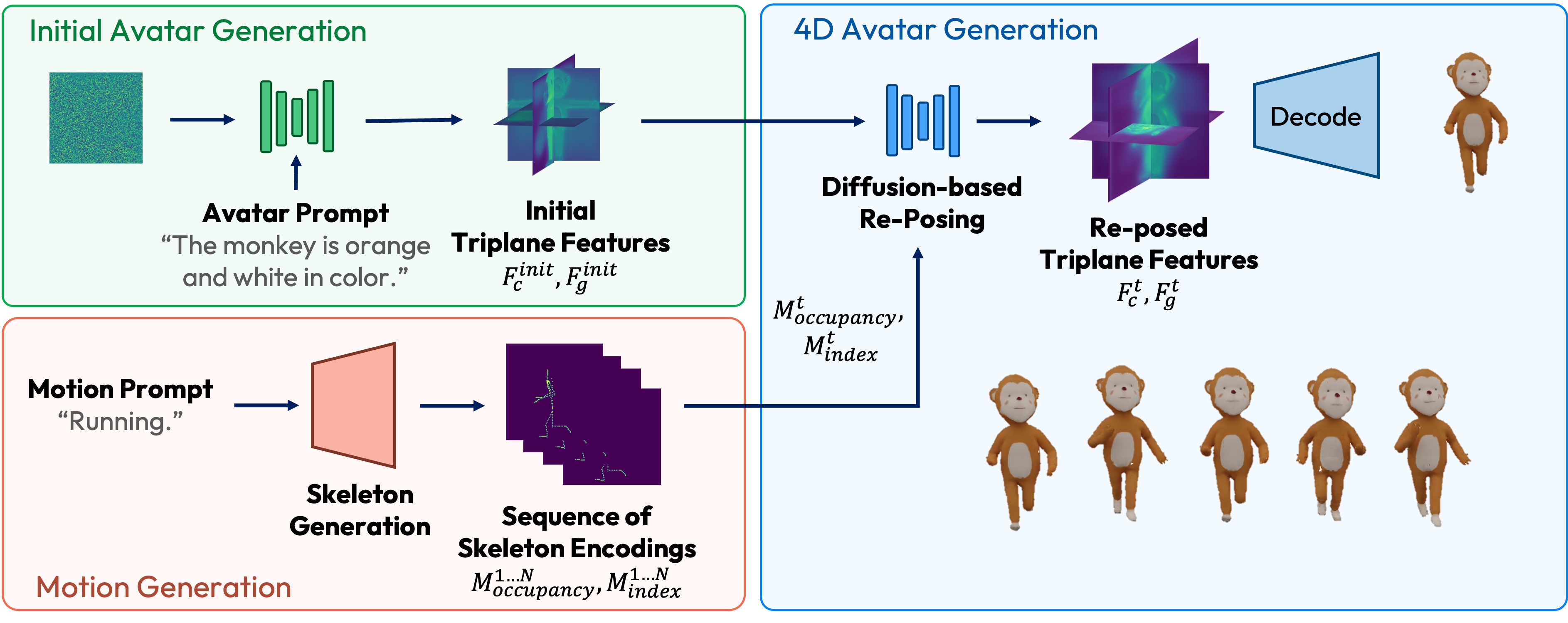}
  \caption{\textbf{Method overview.} Given a prompt, we generate a 4D mesh through a three-step process: (1) \textit{Triplane generation}, which transforms text descriptions into 3D representations; (2) \textit{Skeleton generation}, which generates 3D motion sequences from the text; and (3) \textit{Diffusion-based reposing}, which integrates these components to produce animated 3D avatars with precise pose control.
  }
  \label{fig:pipeline}
\end{figure*}

% \qihao{We should weaken the use of Direct-3D, just call it something like a triplane diffusion model. Same for the motion model}
Our key insight is that 4D avatar generation can be efficiently achieved by explicitly separating 3D structure modeling from motion control, then recombining them through diffusion-based reposing. As illustrated in Fig.~\ref{fig:pipeline}, our \modelname operates through three main stages: we first generate a high-quality static 3D avatar using a triplane diffusion model (Section~\ref{method:init}), simultaneously encode the target motion into a format compatible with our diffusion architecture (Section~\ref{method:skeleton_encode}), and finally unify these components through our novel diffusion-based reposing mechanism (Section~\ref{method:reposing}) that transforms the static avatar according to the motion guidance while preserving appearance consistency throughout the animation sequence.

\subsection{Initial 3D Avatar Generation} \label{method:init}

We employ a triplane-based diffusion model that generates high-quality 3D assets from text prompts. Inspired by DIRECT-3D~\cite{liu2024direct}, we decompose triplane features into separate geometry and color components to facilitate training stability and enable independent optimization of structural and appearance characteristics.

The generated features $T_{\text{init}} = (F_{g}^{\text{init}}, F_{c}^{\text{init}})$ comprise three feature maps corresponding to the $XY$, $XZ$, and $YZ$ coordinate planes, with each plane containing $C$ feature channels. At resolution characterized by $H \times W$, the geometry triplane $F_{g}^{\text{init}} \in \mathbb{R}^{3 \times C \times H \times W}$ encodes structural information while the color triplane $F_{c}^{\text{init}} \in \mathbb{R}^{3 \times C \times H \times W}$ captures appearance details. These triplane features undergo reshaping to $\mathbb{R}^{3 \cdot C \times H \times W}$ for convolutional processing before being decoded through an auto-decoder to generate final 3D renderings. The processed features $T_{\text{init}}$ subsequently function as conditioning inputs for the diffusion-based reposing framework described in Section~\ref{method:reposing}.

\begin{figure*}
    \centering
    \includegraphics[width=0.8\linewidth]{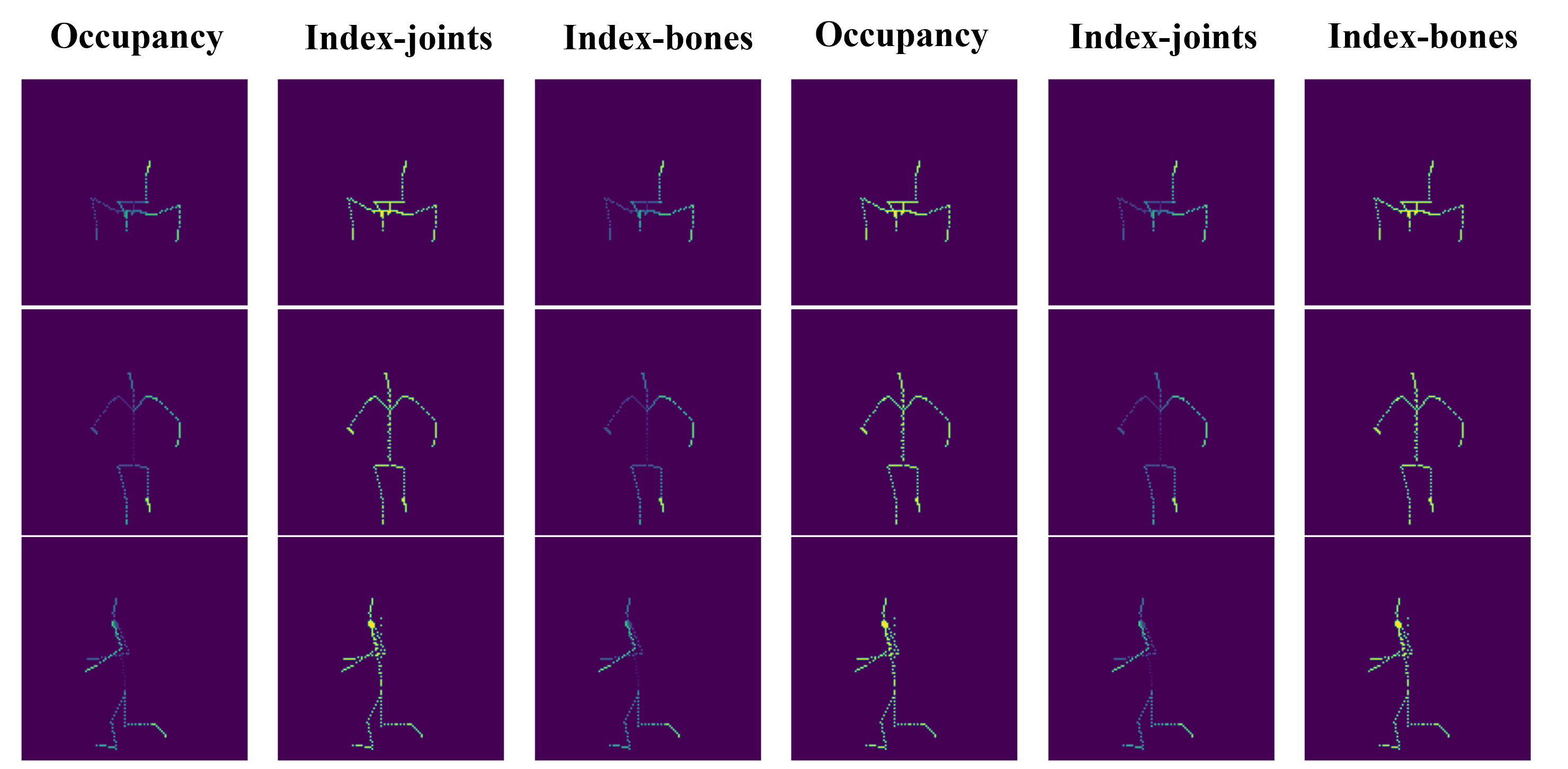}
        \caption{\textbf{Visualization of our triplane skeleton encoding approach.} 
        Each row visualizes the same pose projected onto three orthogonal planes: XY (top), XZ (middle), and YZ (bottom). For each projection, we display three feature channels, repeated to match the triplane feature dimensionality: \textit{occupancy maps}, which highlight the structural presence of joints and bones, and \textit{index maps}, where brightness variations represent normalized joint indices.
}
    \label{fig:encoded_skeleton}
\end{figure*}

\subsection{Skeleton Encoding for Pose Guidance} \label{method:skeleton_encode}
% \todo{add a small intro here}
Our diffusion-based reposing framework requires effective skeleton representations to guide the generation of properly posed triplane features. While 3D skeletons naturally capture the target pose information needed for character animation, their structurally complex nature presents compatibility challenges when used directly with diffusion architectures. This section introduces our skeleton encoding approach that systematically transforms 3D skeletal data into a format specifically designed to condition diffusion models while carefully preserving the spatial relationships essential for accurate triplane feature reposing.

\textbf{Motion Generation.} We integrated a transformer model to generate our motions following works like MoMask \cite{guo2024momask} which demonstrates state-of-the-art performance in synthesizing diverse and realistic movements and implemented a pose transfer pipeline to adapt the motion generation framework to our dataset. The pipeline enables us to work with 3D skeleton that is naturally represented as a set of vertices (joints) and edges (bones), as well as the 3D coordinates of the vertices.
Although this representation encodes rich 3D information, it is not readily compatible with diffusion models, particularly when used as a conditioning signal to guide triplane feature generation.
Hence we propose to pre-process 3D skeletons into 2D triplane skeleton encodings, which enable efficient and effective pose guidance (see Fig.~\ref{fig:encoded_skeleton}).

\textbf{Skeleton encoding.} Given the generated 3D skeleton, we process three 2D skeleton encodings each from a different orthogonal view, \textit{i.e.}, XY, XZ, and YZ. We project the 3D joints to the 2D plane and produce two maps that preserve key skeleton information: (1) the \textit{occupancy map} is a binary map indicating the presence of joints and bones when projected to the 2D plane, and (2) the \textit{index map} encodes the normalized joint index for regions occupied by joints and bones. Specifically, for each location $(u, v)$ in the 2D lattice, the occupancy and index maps are represented by
% \eddie{defination of i j N}
% \begin{align}
% M_\text{occupancy}(u,v) & = \begin{cases}
% 1,\hspace{2.6em} & \text{if $(u,v)$ lies within a joint or bone projection} \\
% 0, & \text{otherwise}
% \end{cases} \nonumber \\
% M_\text{index}(u,v) & = 
% \begin{cases} 
% \frac{i}{N-1}, & \text{if $(u,v)$ is the 2D projection of joint $j$} \\
% \frac{i+j}{2(N-1)}, & \text{if $(u,v)$ lies within the 2D projection of bone $i,j$} \\
% 0, & \text{otherwise}
% \end{cases}
% \end{align}
% \begin{align}
% M_\text{occupancy}(u,v) & = \begin{cases}
% 1, & \text{if $(u,v)$ lies within} \\
%    & \text{a joint or bone projection} \\
% 0, & \text{otherwise}
% \end{cases} \nonumber \\[0.5em]
% M_\text{index}(u,v) & = 
% \begin{cases} 
% \frac{i}{N-1}, & \text{if $(u,v)$ is the 2D} \\
%               & \text{projection of joint $j$} \\
% \frac{i+j}{2(N-1)}, & \text{if $(u,v)$ lies within the} \\
%                     & \text{2D projection of bone $i,j$} \\
% 0, & \text{otherwise}
% \end{cases}
% \end{align}

Although we are representing the 3D skeleton with 2D maps, the 2D skeleton embeddings preserve critical spatial relationships between joints and bones.
As shown in Fig.~\ref{fig:encoded_skeleton}, this approach maintains the structural integrity of the skeleton across multiple orthogonal views (XY, XZ, YZ), ensuring that the complete 3D pose information is retained.

\textbf{Compatibility with triplane diffusion models.} 2D skeleton encoding is naturally compatible with the triplane diffusion models. They can be directly used as conditioning signals or concatenated with noisy latent to guide the generation process. This design avoids the need for an additional encoding module to process the structured 3D data.

\textbf{Computational efficiency.} Converting complex 3D skeleton data into a set of 2D projections significantly reduces computational complexity compared to processing full volumetric representations. This efficiency is critical for real-time applications and allows the diffusion model to process pose information without prohibitive computational demands.
%%%%%%%%%%%%%%%
{\small
\begin{align}
M_\text{occupancy}(u,v) & = \begin{cases}
1, & \text{if $(u,v)$ within joint/bone projection} \\
0, & \text{otherwise}
\end{cases} \nonumber \\
M_\text{index}(u,v) & = 
\begin{cases} 
\frac{i}{N-1}, & \text{if $(u,v)$ is projection of joint $i$} \\
\frac{i+j}{2(N-1)}, & \text{if $(u,v)$ within bone $(i,j)$ projection} \\
0, & \text{otherwise}
\end{cases}
\end{align}
\noindent where $N$ is the total number of joints, $i,j \in \{0,\ldots,N-1\}$ are joint indices, and $(i,j)$ denotes the connecting bone.
}
%%%%%%%%%%%%%%%% 

\subsection{Diffusion-based Avatar Re-posing}\label{method:reposing}

The goal of the re-posing module is to generate re-posed triplane features given initial triplane features $F_\text{geo}^\text{init}$ and $F_\text{color}^\text{init}$ and skeleton representation $S_\text{enc}^t$ at time $t$.
\begin{align}
F_\text{geo}^t, F_\text{color}^t = \text{RePose}\left(F_\text{geo}^\text{init}, F_\text{color}^\text{init}, S_\text{end}^t\right)
\end{align}
We implement the re-posing module as a conditional diffusion model with a U-Net architecture. Specifically, two types of conditioning methods are considered.
% \eddie{explaination of motion gen??}

\textbf{Direct concatenation.} We concatenate the diffusion model latent with the initial triplane features and the skeleton representation, and pass the combined input through a convolutional layer followed by the U-Net backbone for denoising.
This direct integration enables spatial alignment between the avatar features and skeletal structure, providing a strong geometric prior for the denoising process.

\begin{figure*}
    \centering
    \includegraphics[width=\linewidth]{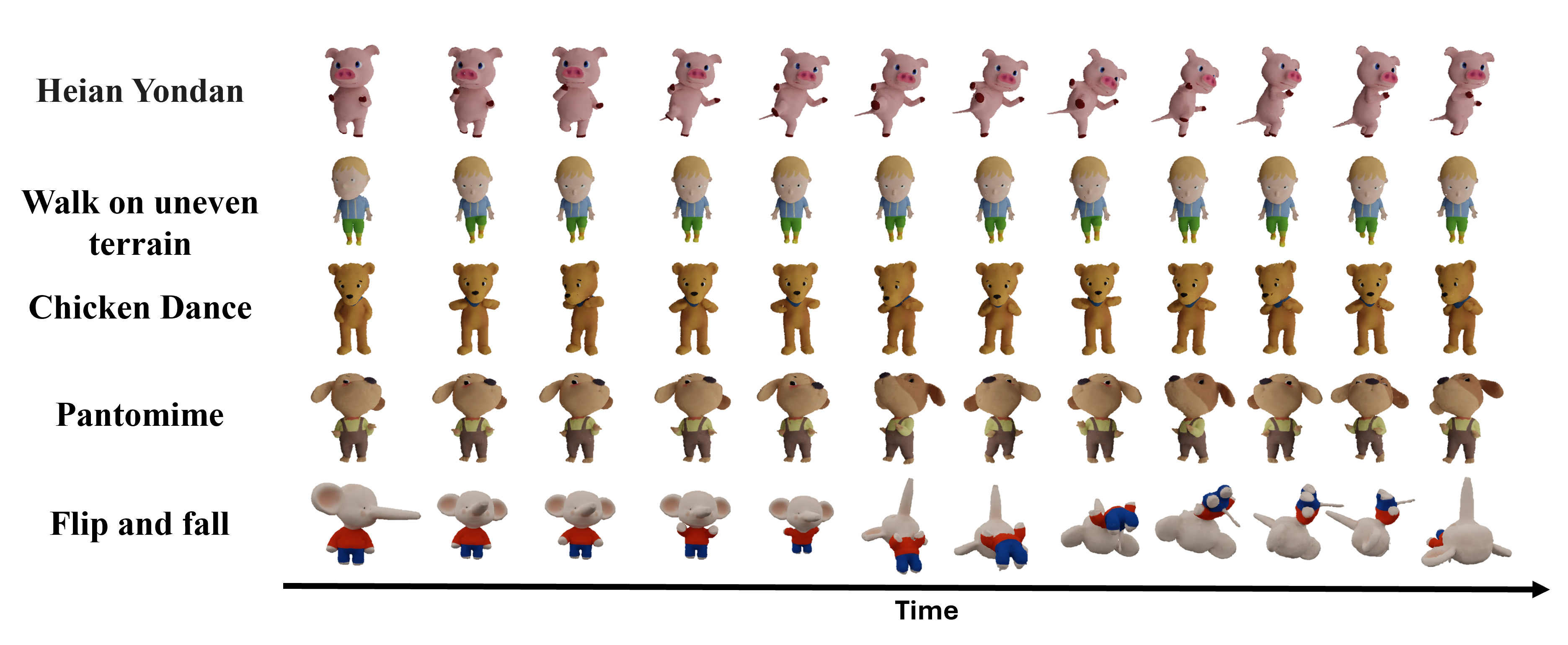}
    \caption{\textbf{Complex motion generation with \modelname.} Our model is capable of generating extreme pose transitions and complex motions while preserving consistent geometry and appearance.}
    % \todo{maybe also add the prompt? Not sure}\eddie{Actually I found it}\qihao{can you guess a close one?LOL}
    \label{fig:back_flip}
\end{figure*}

\textbf{Cross-attention conditioning.} In the U-Net, we convert the image-like condition (triplane appearance and skeleton) onto sequential tokens, allowing the diffusion model to discover meaningful relationships between spatial positions while integrating the conditional information. After processing, it reshapes the enhanced representation back to spatial format and adds it to the original input through a residual connection. It acts as a bridge that precisely maps the target skeleton onto the character features and enables high quality, detailed reposing while preserving the character's original identity and appearance.
% \cite{}
%
These modules use attention mechanisms across multiple resolutions to create dynamic relationships between the input features and conditional information. With this transformer architecture, \modelname achieves superior pose transfer through precise joint-to-feature alignment and reduces reposing artifacts such as limb distortion and geometry collapse during extreme poses \cite{ren2023dreamgaussian4d}, and result shown in Fig.~\ref{fig:back_flip}.

\section{Experiments}
\subsection{Experimental Setup}
\noindent\textbf{Datsets.}\label{experimental_setup}
Given the scarcity of comprehensive 4D datasets, we strategically selected and combined two complementary resources to address our research needs.
For character representation, we utilized the RaBit dataset~\cite{Luo_2023_CVPR}, which comprises 1,500 unique character models including both human and anthropomorphic animal figures. This dataset provides SMPL-like parametric models that enable precise manipulation of pose through joint rotation and skeleton configurations. This structure allows for consistent deformation across diverse character morphologies.

To ensure robust motion diversity and anatomically realistic movement patterns, we incorporated the AMASS dataset~\cite{AMASS:ICCV:2019}~\cite{Guo_2022_CVPR}, which offer an extensive collection of high-fidelity human motion sequences captured from real subjects. These datasets provided the necessary variation in movement dynamics to train our models while maintaining biomechanical plausibility across different action categories. Our approach is specifically designed to be compatible with existing motion generation frameworks without requiring additional training, enabling efficient implementation with minimal computational resources. This compatibility significantly reduces the operational costs associated with deploying advanced motion synthesis systems in production environments.

\noindent\textbf{Evaluation.} We evaluate our method against state-of-the-art approaches for 4D avatar generation and animation, focusing on both computational efficiency and animation quality. Our baseline comparisons include optimization-based methods such as MAV3D \cite{singer2023text4d}, Consistent4D \cite{jiang2024consistentd}, and DreamGaussian4D \cite{ren2023dreamgaussian4d}, which represent diverse approaches to 4D content creation. We also measure perceptual quality (LPIPS), semantic alignment (CLIP), and temporal coherence (FVD) using the Consistent4D benchmark in Tab. \ref{tab:quantitative_results}.
\begin{figure*}[h]
    \centering
    \includegraphics[width=0.85\textwidth]{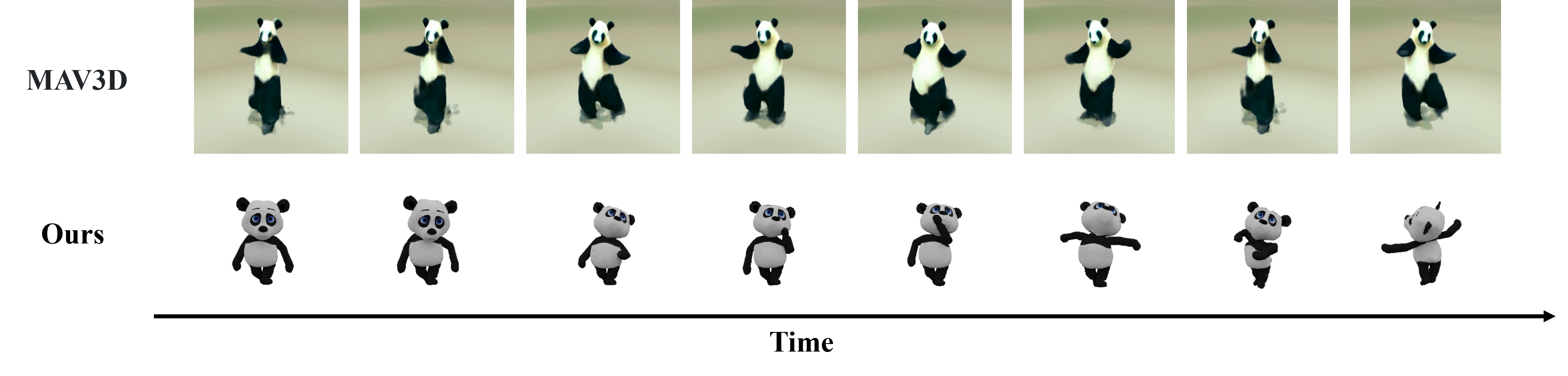}
    \vspace{-3mm}
    \caption{\textbf{Comparison with MAV3D.} 
    Our method generates more realistic characters with improved visual quality, enhanced pose articulation, and consistent appearance throughout the animation sequence.
    Baseline results are obtained from the official project page.
    % \todo{Do I need more images like \ref{fig:back_flip}?}\qihao{Yes, it is better to have more results.}
    }
    \label{fig:enter-label}
\end{figure*}

\begin{figure*}[t]
    \centering
    \includegraphics[width=0.85\linewidth]{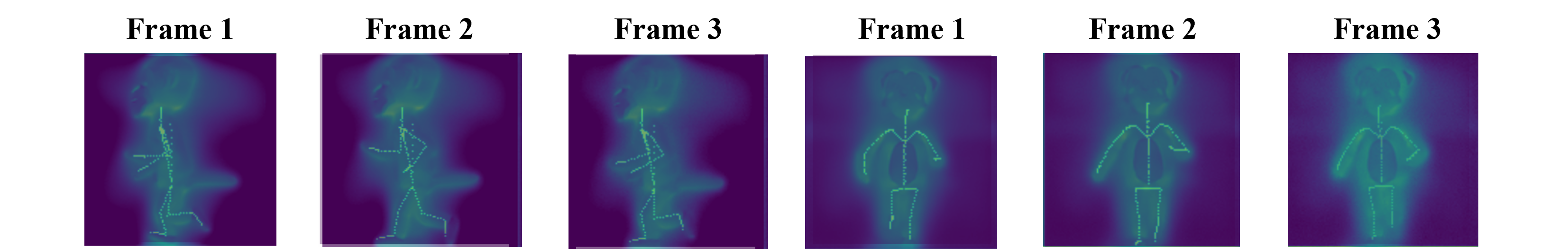}
    % \vspace{0.15cm}
    \caption{\textbf{Triplane feature visualization.} 
    These frames illustrate the integration of the skeleton features with avatar appearance features, projected onto the same plane across sequential time steps. The visualization shows that the generated features align precisely with the target skeletal structure, confirming that our approach effectively produces geometry that adheres to skeletal guidance while preserving consistent appearance attributes.
    }
    % \todo{Put this figure in one row}
    \label{fig:triplane_feat_frame}
\end{figure*}
\begin{table}[t]
    \centering
    \caption{\textbf{Inference speed and iteration requirements.} 
    Our method enables significantly faster generation without iterative optimization.}
    \resizebox{\linewidth}{!}{%
    \begin{tabular}{lcc}
    \hline
    Method & Time & Iterations \\
    \hline
    MAV3D \cite{singer2023text4d} & 6.5 hr & 12k \\
    Animate124 \cite{zhao2023animate124} & - & 20k \\
    Consistent4D \cite{jiang2024consistentd} & 2.5 hr & 10k \\
    4D-fy \cite{bahmani20244d} & 23 hr & 120k \\
    Dream-in-4D \cite{zheng2024unified} & 10.5 hr & 20k \\
    AYG \cite{ling2024align} & - & 20k \\
    DG4D (Image-to-4D GS) \cite{ren2023dreamgaussian4d} & 6.5 mins & 0.7k \\
    DG4D + Texture Refinement & 10 mins & 0.75k \\
    \hline
    \modelname (Ours) & \textbf{0.6 mins} & \textbf{1}$^\star$ \\
    \hline
    \end{tabular}}
    \vspace{2mm}
    {\footnotesize $^\star$ Single forward pass, no optimization required.}
    \label{tab:generation_time}
\end{table}
\begin{table}[t]
    \centering
    \caption{Quantitative comparison on the Consistent4D benchmark.}
    \label{tab:quantitative_results}
    \begin{tabular}{lccc}
    \toprule
    Method & LPIPS$\downarrow$ & CLIP$\uparrow$ & FVD$\downarrow$ \\
    \midrule
    Consistent4D & 0.16 & 0.87 & 1133.44 \\
    4DGen & 0.13 & 0.89 & - \\
    STAG4D & 0.13 & 0.91 & 992.21 \\
    DG4D~ & 0.16 & 0.87 & - \\
    Efficient4D~\cite{pan2025efficient4dfastdynamic3d} & 0.14 & 0.92 & - \\
    L4GM & 0.12 & 0.94 & 691.87 \\
    \midrule
    \textbf{TriDiff-4D (Ours)} & \textbf{0.13} & \textbf{0.94} & \textbf{626.29} \\
    \bottomrule
    \end{tabular}
\end{table}

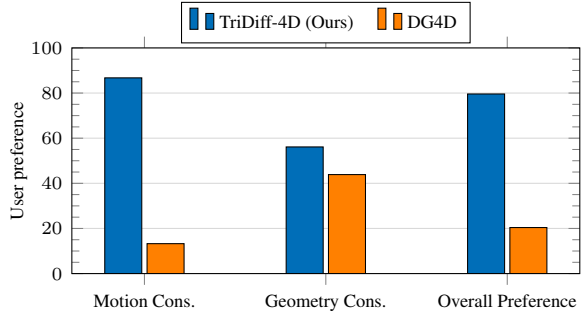
\begin{figure}[t]
    \centering
    \begin{tikzpicture}
      \begin{axis}[
        ybar,
        bar width=14pt,
        ymajorgrids=true,
        grid style={draw=gray!30},
        width=\linewidth,
        height=0.55\linewidth,
        enlarge x limits=0.2,
        ylabel style={font=\scriptsize, yshift=-15pt},
        ylabel={User preference},
        ymin=0, ymax=100,
        ytick distance=20,
        minor y tick num=3,
        tick label style={font=\scriptsize},
        symbolic x coords={Motion Cons., Geometry Cons., Overall Preference},
        xtick=data,
        legend style={
            at={(0.5,1.02)},
            anchor=south,
            legend columns=-1,
            font=\scriptsize
        },
      ]
        \addplot[fill=NavyBlue] coordinates {
          (Motion Cons.,86.73)
          (Geometry Cons.,56.12)
          (Overall Preference,79.59)
        };
        \addplot[fill=orange] coordinates {
          (Motion Cons.,13.26)
          (Geometry Cons.,43.87)
          (Overall Preference,20.40)
        };
        \legend{\modelname\ (Ours)$\quad$, DG4D}
      \end{axis}
    \end{tikzpicture}
    \caption{\textbf{User preference study results comparing \modelname with DreamGaussian4D.} 
    Participants consistently preferred \modelname across all evaluation criteria.}
    \label{fig:user_preference}
\end{figure}

\noindent\textbf{Implementation details.} We use the Adam optimizer~\cite{kingma2014adam} with a learning rate of $1e^{-4}$ for the diffusion model. The learning rate schedule includes a linear warm-up over the first 500 iterations and a step decay that reduced the rate by half after 500K iterations. Our model operates on 128 $\times$ 128 resolution inputs and is trained for 2M iterations on a dataset comprising 1.5K characters, each with 22 poses which contains total of 33K training samples.
More experimental details in supplementary

\subsection{State-of-the-Art Text-to-4D Generation} \label{state_of_the_art}
% \todo{@qihao, add user preference results}

We compare our model with other 4D generation models in Tab.~\ref{tab:generation_time}, which demonstrates the substantial computational efficiency of our approach. Our method generates a complete 14 frame animation sequence in merely 0.6 minutes, representing a significant reduction in processing time compared to existing techniques that require hours of computation. This efficiency stems from our non-iterative pipeline design, which requires only single pass rather than thousands of optimization iterations. 

Beyond computation advantages, our approach maintains superior geometric consistency during animation shown in Fig.~\ref{fig:Jelly_effect} that preserves structural integrity during pose changes, whereas alternative approaches have a jelly-like wobbling effect. The effect is enabled by our triplane representation (Fig.~\ref{fig:triplane_feat_frame}) which maintains consistent feature information across frames.

Finally, we conduct a user preference study the results from DreamGaussian4D and \modelname, where each generated animation was reviewed by 14 participants. Participants are asked to select the best one in three evaluation axes: geometry consistency, motion consistency and overall preference. The results are presented in Tab.~\ref{fig:user_preference}, where \modelname demonstrates a clear advantage across all evaluation criteria, with users strongly preferring our approach for its superior visual quality and motion coherence.

\subsection{Ablation Studies}
\begin{figure}
    \centering
    \includegraphics[width=\linewidth]{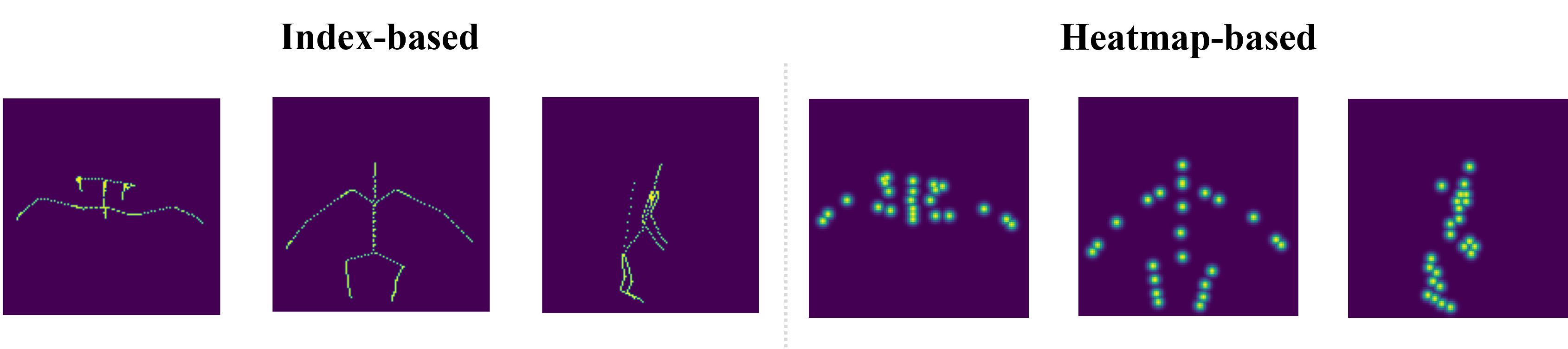}
    \caption{\textbf{Different skeleton encoding approaches.} Left: Index-based skeleton encoding, where joints are represented as discrete points across three orthogonal planes. Right: The alternative heatmap-based encoding that provides a smoother, continuous representation of joint locations.}
    \label{fig:heatmap}
\end{figure}
% \todo{@eddie, waiting for results}
To evaluate the impact of key design decisions in our \modelname framework, we conduct a series of ablation experiments to examine both the skeleton encoding method and attention resolution. 

\begin{figure}
    \centering
    \vspace{-3mm}
    \includegraphics[width=\linewidth]{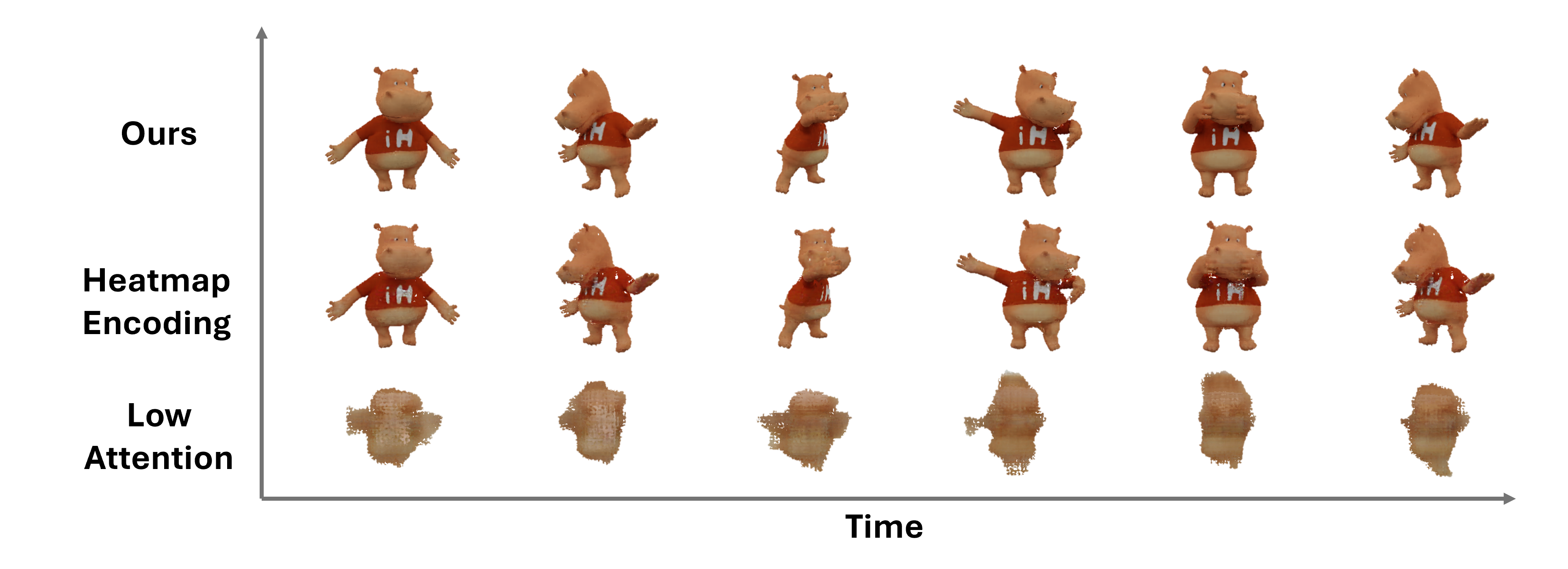}
    \caption{\textbf{Visualization of ablation studies.} (1) Top: Index-based skeleton full configuration. (2) Mid: Heatmap-based skeleton with full configuration. (3) Bottom: Index-based skeleton with lower attention resolution and head number.}
    \label{fig:alblationstudies}
\end{figure}

\textbf{Skeleton Encoding Method.} We compare our index-based skeleton encoding method with a heatmap-based approach (Fig.~\ref{fig:heatmap}). The fundamental difference between these approaches lies in how skeleton information is represented spatially. Our index-based method provides precise skeletal structures Fig.~\ref{fig:triplane_feat_frame} through normalized joint indices and connections, which creates a sharp localization of object skeleton structure. In contrast, the heatmap-based approach diminishes the connection information between joints, resulting in less defined skeleton constraints during the diffusion process.
Experimental results demonstrate that our index-based approach achieves better pose accuracy and superior appearance quality. While using the heatmap-based encoding, the diffusion model struggles to maintain structural integrity during complex animations, leading to noticeable visual artifacts including holes in the character's body surface. 

\textbf{Attention Mechanism.} We reduce the attention resolution from (32, 16, 8) to (4, 2, 1) and decrease attention heads from 4 to 1. Despite a 5\% reduction in memory usage, this configuration produces significantly degraded results (Fig.~\ref{fig:alblationstudies}). With the same training iterations, the lower attention model fails to learn proper reposing patterns, resulting in severe geometry distortions and loss of character details. These results confirm that maintaining sufficient attention resolution is critical for high-fidelity animation generation, demonstrating that our configuration strikes an optimal balance between quality and efficiency.
Despite the theoretical computational advantage of lower resolution attention, we observe that training with this configuration is actually slower in practice. This result occurs because the reduction of spatial resolution prevents the model from efficiently extracting and focusing on the most relevant features for pose transfer. The lower resolution attention maps become too coarse to capture important local relationships between triplane skeleton structure and geometry. Furthermore, the reduced number of attention heads severely limits the model's ability to simultaneously attend to different aspects of the animation process, such as maintaining overall character structure while positioning individual limbs.
% \eddie{Reviewer says we do not have CLIP Score or VisionReward for benchmark. Do we need it? Because of existing performance performance for Direct3D and momask. Maybe explain using the skeleton and 3D model is super aligned and not need other scores}

\section{Conclusion}
In this paper, we present \modelname, a novel approach for generating high-quality 4D avatars through diffusion-based triplane reposing. Our method addresses critical limitations of existing 4D generation techniques, regarding temporal consistency, motion accuracy, visual fidelity, and computational efficiency combining a specialized diffusion model with the triplane representation and skeleton-guided conditioning. Notably, our approach achieves precise skeleton-based pose control while reducing generation time, enabling significantly faster workflow and real-time applications previously unattainable with existing methods.

\clearpage

\appendix

% For two-column format, use \newpage instead of \clearpage
\newpage

% Remove the manual page counter reset - not needed for two-column
% Keep the section formatting
\section*{Appendix}
\setcounter{section}{0}
\renewcommand{\thesection}{\Alph{section}}

\section{Implementation Details}  
\subsection{Initial Static Avatar Generation} 
Our dataset comprises 1,500 unique characters, each rendered in multiple poses to create a diverse training set of 3,000 initial configurations. Through extensive experimentation, we discovered that initializing with a T-pose consistently yields superior performance and faster training convergence. While our model can process avatars in any initial pose, non-T-pose initializations generally require additional training iterations to achieve comparable reposing quality.

\subsection{Training} 
We employ a two-stage training approach. The static avatar generation model is trained for 200,000 iterations, while the reposing model undergoes a more extensive 2M iterations on 33K poses derived from our character dataset. Both models are trained on 4 NVIDIA H100 GPUs.

For the triplane feature representation, we set the number of channels C = 6 and train the diffusion model with 1,000 diffusion steps using a linear noise schedule. To optimize efficiency, we implement a progressive reconstruction strategy:

\noindent\textbf{For the initial avatar model:}
\begin{itemize}
    \item 15 reconstructions per iteration for the first 50,000 iterations
    \item 3 reconstructions per iteration until 100,000 iterations
    \item 1 reconstruction per iteration for all remaining iterations
\end{itemize}

\noindent\textbf{For the reposing model:}
\begin{itemize}
    \item 30 reconstructions per iteration for the first 50,000 iterations
    \item 15 reconstructions per iteration until 300,000 iterations
    \item 5 reconstructions per iteration until 800,000 iterations
    \item 1 reconstruction per iteration for the final training phase
\end{itemize}

This progressive schedule significantly accelerates training while maintaining output quality by allocating more computational resources during early learning phases when the model requires more intensive refinement.

\section{Limitations} 
Our current model does not simulate cloth dynamics and only generate human avatars due to the scarcity of appropriate 4D datasets that include realistic cloth behavior. This limitation stems directly from the broader challenge in the field: the lack of comprehensive training data that captures how garments fold, wrinkle, and respond to character movement during animation. Without such specialized datasets, our model cannot learn the complex physical interactions between clothing and body motion, particularly evident in characters with loose or flowing garments where these dynamics are most noticeable. 

Moreover, our framework presently employs standard diffusion models. The exploration of more advanced approaches, such as Flow Matching~\cite{albergo2022building, lipman2022flow, liu2025flowing, liu2022flow} and enhanced architectures~\cite{peebles2023scalable, liu2024alleviating}, is left to future work. 

\section{Ethics and Broader Impact}  
Our 4D avatar generation research raises ethical considerations that we actively address. While our technology focuses on non-photorealistic characters rather than realistic human representations, we strongly condemn any use of our technology to create deceptive content intended to harm individuals, groups, or organizations, or to deliberately spread misinformation in public discourse.

Our research on diffusion-based reposing has potential applications beyond graphics research. The ability to rapidly transform avatar poses through feature-space modifications could enhance virtual reality experiences, personalized avatars and real-time applications where traditional optimization approaches are prohibitively slow.

\section{Safeguards} 
Given the capabilities of \modelname, it is essential to establish appropriate safeguards against potential misuse. Our training dataset deliberately excludes photorealistic human characters, focusing instead on stylized figures, which inherently limits the model's ability to generate deceptive content that could be used for harmful purposes.

% \section{New Assets} 
% For this research, we created a specialized 4D dataset by combining characters from the RaBit \cite{Luo_2023_CVPR} dataset with motion sequences from AMASS \cite{AMASS:ICCV:2019}. The RaBit dataset contains 1,500 different characters that can be parametrically modified to generate additional variations, allowing us to significantly expand our training data through controlled parameter manipulation. By applying diverse motion sequences from AMASS to these parametric models, we created a comprehensive training dataset of 33K samples capturing both varied character morphologies and realistic animations. 

% Additionally, we developed custom training code for our diffusion-based triplane reposing approach, including our novel skeleton encoding methods and conditional diffusion architectures. These assets will be made publicly available upon publication to facilitate reproducibility and further research in 4D character animation.

\clearpage
% Fixed figure formatting for two-column
\begin{figure*}
    \centering
    \includegraphics[width=\textwidth]{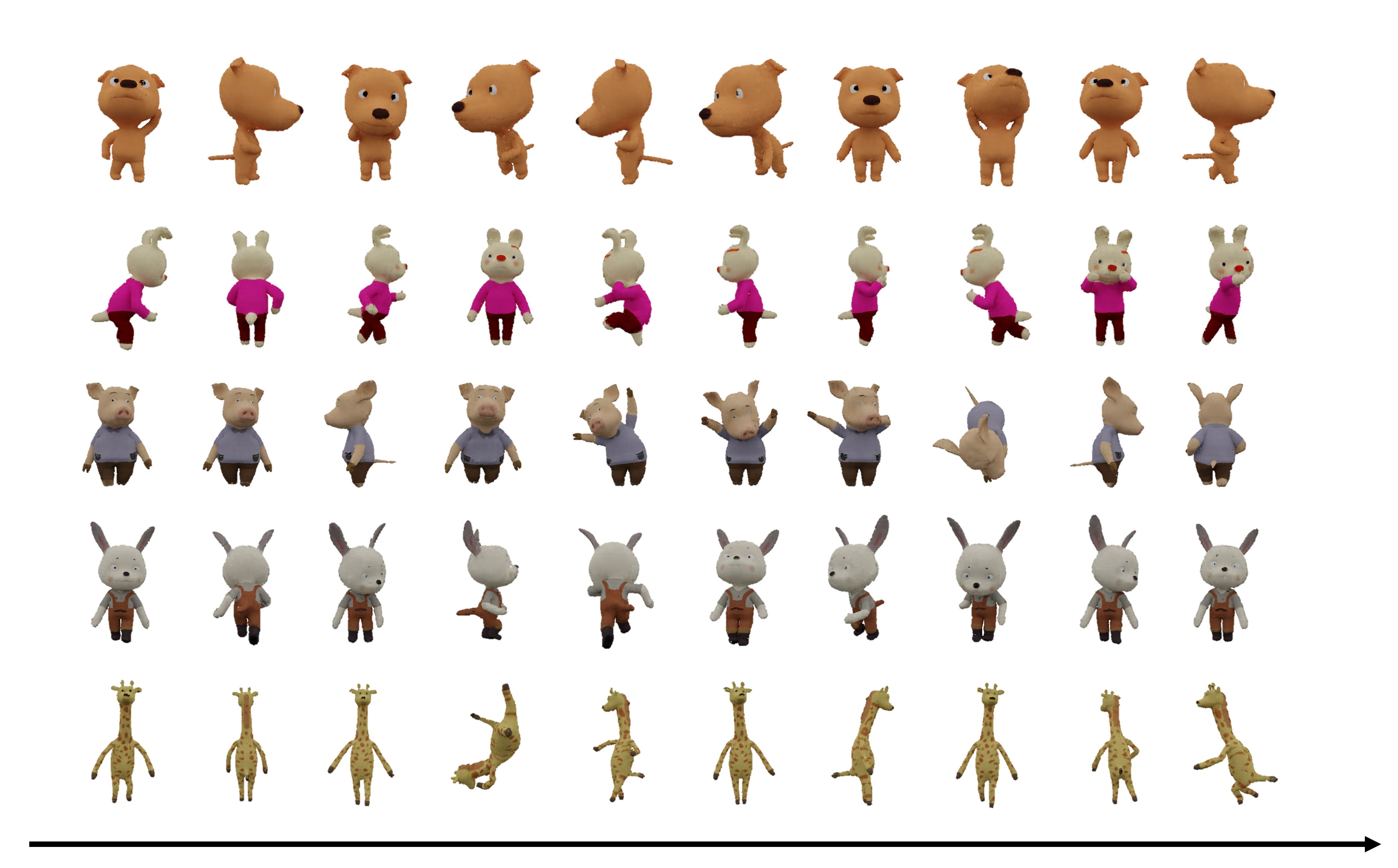}
    \caption{Additional results using \modelname}
    \label{fig:supp_img}
\end{figure*}

\clearpage
{
    \small
    \bibliographystyle{ieeenat_fullname}
    \bibliography{main}
}

\end{document}